\documentclass{article}

\usepackage{spconf,amsmath,amssymb,upgreek,graphicx,booktabs,array,xcolor}
\usepackage{hyperref}
\usepackage{microtype}
\usepackage{flushend}

\hypersetup{hidelinks}
\makeatletter
\renewcommand{\name}[1]{\gdef\@name{{\em #1}}}
\def\@maketitle{\newpage
 \null
 \vskip 2em \begin{center}
 {\large \bf \@title \par} \vskip 1.5em {\large \lineskip .5em
\begin{tabular}[t]{c}\@name \\ \@address
 \end{tabular}\par} \end{center}
 \par
 \vskip 1.5em}
\renewcommand\section{\@startsection{section}{1}{\z@}%
  {-1.7ex plus -.5ex minus -.2ex}{.7ex plus .2ex}{\normalfont\large\bfseries}}
\renewcommand\subsection{\@startsection{subsection}{2}{\z@}%
  {-1.3ex plus -.4ex minus -.2ex}{.5ex plus .2ex}{\normalfont\normalsize\bfseries}}
\let\standardthebibliography\thebibliography
\renewcommand{\thebibliography}[1]{%
  \standardthebibliography{#1}%
  \small
  \setlength{\itemsep}{-2pt}%
  \setlength{\parsep}{0pt}%
}
\makeatother

\newcommand{\tauvoice}{\texorpdfstring{$\uptau$}{tau}-Voice}
\newcommand{\passone}{\mbox{Pass@1}}
\newcommand{\passk}{\texorpdfstring{$\mathrm{Pass}^{3}$}{Pass3}}
\newcommand{\telicit}{\texorpdfstring{$\uptau$}{tau}-Elicitation}

\title{TAU-ELICITATION: BENCHMARKING MULTI-TURN ENTITY EXTRACTION IN VOICE AGENTS}

\name{Soham Ray$^{1}$ \qquad Victor Barres$^{2}$}
\address{$^{1}$Sierra \qquad $^{2}$Mercor \\
{\small\texttt{soham@sierra.ai}}}

\begin{document}

\maketitle

\begin{abstract}
Voice agents often need to collect names, addresses, identifiers, dates, and
times exactly, yet end-to-end benchmarks obscure where capture fails. We
introduce \telicit{}, a 200-task voice benchmark spanning 10 entity types,
controlled difficulty, caller realisms, and three environments. A matched text
agent passes all tasks, but four voice configurations achieve robust exact
success on only 0.14--0.41. Agents increase verification for hard and
unfamiliar entities and sometimes for incorrect captures, but not for their
weakest caller voice; only 24--37\% of verified errors are repaired. A scaffold
that prescribes spelling, read-back, correction, and confirmation raises
robust \passk{} by 14--31 points, at a cost of 21--28 seconds per call.
Realisms such as spelling variations and restarts do not detectably affect
exact success; mispronunciation increases repair effort. These results
identify strategy selection and successful recovery as the central bottlenecks
in exact spoken entity collection.
\end{abstract}

\begin{keywords}
voice agents, spoken dialogue evaluation, multi-turn entity extraction,
conversational repair, task-oriented dialogue
\end{keywords}

\section{Introduction}
\label{sec:introduction}

Voice-agent benchmarks usually score complete tasks
\cite{ray2026tauvoice,bogavelli2026evabench,bhosale2026duplexworld}, yet
authentication and key-entity transcription remain common failure points.
Existing work does not isolate whether an interactive agent can recognize an
uncertain capture, decide when to slow down, repair an error, and store the
exact value. Nor does it measure how this capability changes with entity type,
difficulty, repeated conditions, or caller effort. This matters because
misunderstood named entities impede spoken dialogue, while confirming every
attribute can itself frustrate callers \cite{filisko-seneff-2004-error}.

\telicit{} tests this capability on short callbacks about missing database
fields. Each core task elicits one missing value, isolating the atomic unit of
spoken entity collection and establishing a minimum-complexity floor for
broader calls; linked multi-entity tasks then test how errors compound. Its default
prompt defines the \emph{agent-directed} condition: the record must be exactly
correct and letter-by-letter verification is suggested when unsure, but the
agent decides when and how to verify. A matched \emph{scaffolded} condition
instead prescribes asking, spelling or reading back, confirming, correcting,
and then submitting. Both share consent, never-guess, logging, and
one-submission rules, isolating strategy selection from protocol execution.
Deterministic database comparison scores both conditions without an LLM judge.

We contribute a reproducible 200-task benchmark spanning 10 entity types,
controlled difficulty and caller realisms, and multi-entity calls. It measures
capture, verification, repair, effort, and robust success across three
environments; \href{https://github.com/sierra-research/tau2-bench/blob/tau-elicitation/papers/tau-intake/v1/reproduction/README.md}{\textcolor{blue}{code and artifacts are available on GitHub}}.

\begin{figure*}[t]
\centering
\includegraphics[width=\textwidth]{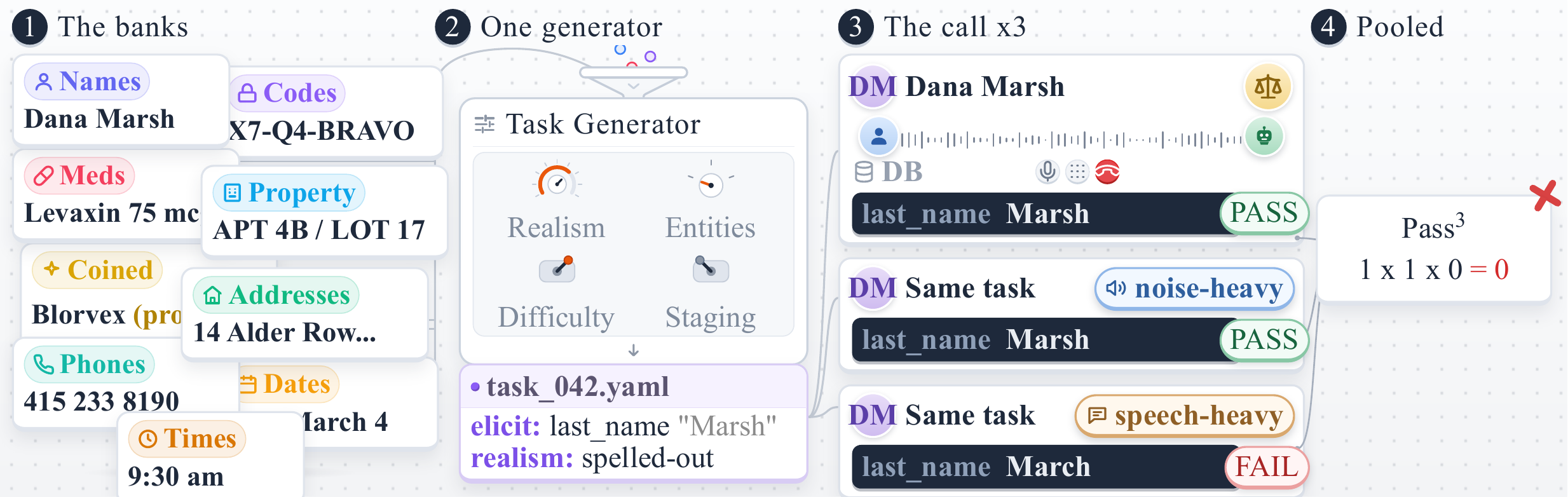}
\caption{\telicit{} selects values from entity banks, builds a controlled task,
and runs it through the \tauvoice{} interaction loop. \passk{} requires success
in all three environment realizations. The scale marks fidelity-judged agent
speech.}
\label{fig:pipeline}
\end{figure*}

\section{Related Work}
\label{sec:related-work}

\begin{table}[t]
\caption{Coverage of related benchmarks. Task: executed tool task; Exact: exact entity score;
$N$: controlled entity count; D: diagnostic or partial coverage.}
\label{tab:related}
\centering
\small
\setlength{\tabcolsep}{2pt}
\renewcommand{\arraystretch}{.9}
\begin{tabular}{@{}lccccc@{}}
\toprule
Work & Multi & Task & Exact & $N$ & Repair \\
\midrule
Voice agents \cite{ray2026tauvoice,bogavelli2026evabench,bhosale2026duplexworld,lin2026fdbv3} & $\checkmark$ & $\checkmark$ & D & -- & D \\
Spoken dialogue \cite{si2023spokenwoz,zhang2022slot} & $\checkmark$ & -- & $\checkmark$ & -- & -- \\
Speech NER \cite{shon2022slue,yu2024heard,assemblyai2026mer} & -- & -- & $\checkmark$ & -- & -- \\
\telicit{} & $\checkmark$ & $\checkmark$ & $\checkmark$ & $\checkmark$ & $\checkmark$ \\
\bottomrule
\end{tabular}
\end{table}

Existing voice-agent benchmarks---\tauvoice{}, EVA-Bench, DuplexWorld, and
Full-Duplex-Bench-v3---measure end-to-end task performance and interaction
quality; several also exercise tool use
\cite{ray2026tauvoice,bogavelli2026evabench,bhosale2026duplexworld,lin2026fdbv3}.
Their broad scope reveals failures such as authentication---identified as the
dominant bottleneck in \tauvoice{} and reported separately in EVA-Bench---but
does not isolate exact entity capture
\cite{ray2026tauvoice,bogavelli2026evabench}. \telicit{} extends the
\mbox{$\uptau$-bench}, \mbox{$\uptau^2$-Bench}, and \tauvoice{} lineage by making
controlled voice entity collection the primary task
\cite{yao2024taubench,barres2025tau2bench,ray2026tauvoice}.

Dialogue-state tracking spans MultiWOZ, SpokenWOZ, and cross-utterance sub-slot
tracking
\cite{budzianowski2018multiwoz,si2023spokenwoz,zhang2022slot}.
ASR and spoken-entity work covers natural-speech NER, ASR error propagation,
unseen entities in synthesized audio, and industrial missed-entity rates
\cite{shon2022slue,szymanski2023ner,yu2024heard,assemblyai2026mer}.
\emph{Error Detection and Recovery in Spoken Dialogue Systems} studies repair strategies
\cite{filisko-seneff-2004-error}. None combines live tool execution, exact
database writes, controlled entity type, difficulty and count, an agent's
choice to verify. Table~\ref{tab:related} summarizes this gap.

\section{Methods}
\label{sec:methods}

\subsection{Task and elicitation policy}
\label{sec:elicitation-policy}

Each task is a callback about a record with missing fields. The agent sees the
field names, asks the caller for their values, logs its captures, and submits
once; the caller reveals a value only after it is requested. The simulated
caller is cooperative: it supplies requested values, spells accurately when
asked, and confirms or corrects read-backs, but never volunteers information.

\subsection{Values, difficulty, and realisms}

Ten banks each contain 40 easy and 40 hard values; the benchmark samples 10 of
each difficulty per bank. Difficulty is type-relative: for example, hard names
are rarer, hard codes are longer or confusable, and hard emails contain more
symbols. Table~\ref{tab:entity-results} reports bank construction and
performance.

A seeded catalog adds six caller realisms by appending a scripted instruction
to the caller prompt. For example, one spelling-style instruction tells the
caller to say ``zed'' for the letter Z whenever spelling a value.
Self-correction supplies a wrong value and immediately repairs it; spelling
variation changes how letters or digits are grouped; falter-and-restart
spelling introduces a mid-string correction; partial and wrong-field answers
require a follow-up; and reviewed mispronunciations alter only the audio.
Conditional behaviors occur only when the conversation creates an opportunity,
so a spelling correction, for example, cannot fire until spelling begins. Each
task stores its generator and catalog versions, seed, selected realism, and
expected database state. The generator also emits linked two- and three-field
tasks, both as flat calls and as a staged workflow that validates one field
before revealing the next.

\subsection{Scoring and behavioral measures}

We compare each final database value with the exact gold string and report
regular-realization success as \passone{}. We define robust \passk{} as exact
success on the same task across three prespecified environment realizations:
\emph{regular};
\emph{noise-heavy}, with 10\,dB background noise, bursts, frame drops, and
muffling; and \emph{speech-heavy}, with interruptions, backchannels, and
restarts. Unlike a repeated-trial score, \passk{} crosses these prespecified
conditions; voices and realism assignments are also redrawn. It therefore asks
whether the same task succeeds robustly, rather than averaging away a failure
in one condition. A 700-call audit placed the 90th-percentile duration at 1.89
minutes, so we use a conservative four-minute limit. Timed-out calls remain in
the denominator and score zero.

We score both the first captured value and the final database value against the
gold string. Silent caller tools log every spelling request and read-back; their
sum is our measure of verification effort. We compare this effort across noise,
entity difficulty and familiarity, and each system's best- and worst-performing
caller voices.

\section{Experiments}
\label{sec:experiments}

\noindent\textbf{Systems.}
We evaluate gpt-realtime-2 (minimal and xhigh reasoning),
gemini-3.1-flash-live-preview (high), and grok-voice-think-fast-1.0
(provider default)
\cite{openai2026models,google2026models,xai2026grokvoicethinkfast}.
The caller uses gpt-5.5 (xhigh, temperature zero) \cite{openai2026models},
with speech synthesized by ElevenLabs eleven\_v3
\cite{elevenlabs2026elevenv3}.

\smallskip
\noindent\textbf{Matched runs.}
Each configuration runs all 200 tasks under the regular, noise-heavy, and
speech-heavy conditions. We repeat this grid with the scaffolded prompt, which
prescribes spelling and read-back behavior. Within each condition, systems use
the same tasks and random seed. This gives 600 agent-directed and 600
scaffolded calls per configuration, or 4,800 calls in total.

\smallskip
\noindent\textbf{Uncertainty and significance.}
Confidence intervals come from 10,000 task-level bootstrap samples. When a task
is sampled, its three linked conditions stay together. Pairwise system tests
also align calls by task. We use a two-sided sign-permutation test with 100,000
permutations and the standard $+1$ correction. Holm correction is applied
separately to the six system comparisons for \passone{} and \passk{}.

\smallskip
\noindent\textbf{Human and judge validation.}
Two raters independently review 90 failed calls across providers, labeling
failure source and subtype. After discussion and six explicit adjudications,
failure source is resolved for 86 of 90 calls and subtype for 66 of the 81
calls resolved as agent failures.
For fidelity validation, gemini-3.1-pro-preview (temperature zero) scores a
frozen stratified 60-utterance sample enriched for human flags
\cite{google2026models}.

\section{Results}
\label{sec:results}

\begin{figure}[t]
\centering
\includegraphics[width=\columnwidth]{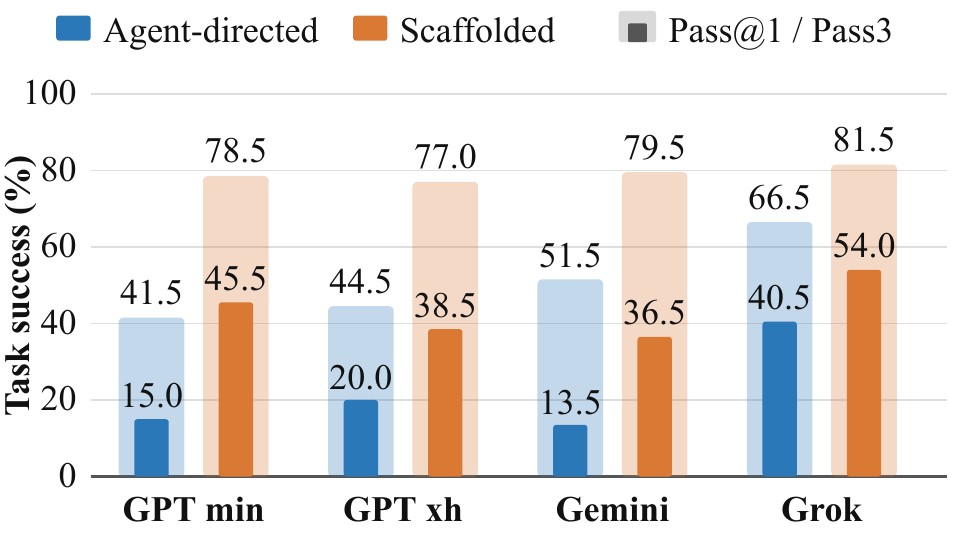}
\caption{Task success over the combined task set. Each pair shows
agent-directed and scaffolded runs; light full bars show \passone{},
and dark inset bars show robust \passk{}.}
\label{fig:main-results}
\end{figure}

\subsection{Strategy selection and robustness}

\begin{list}{\textbullet}{
\setlength{\leftmargin}{1.15em}
\setlength{\labelwidth}{0.45em}
\setlength{\labelsep}{0.35em}
\setlength{\itemsep}{0pt}
\setlength{\parsep}{0pt}
\setlength{\parskip}{0pt}
\setlength{\topsep}{0pt}
}
\item \textit{Scaffolding trades time for reliability.} It raises \passone{}
by 15--37 points and \passk{} by 14--31 points
(Figure~\ref{fig:main-results}), while adding 21--28 seconds per call. Gains are
slightly larger for hard entities (27 versus 23 points).
\item \textit{A supplied protocol narrows provider gaps.} Grok remains best
and significantly exceeds every other agent-directed system on both measures
(Holm-adjusted $p\leq.0032$); no other pair differs. Systems execute a supplied
protocol more reliably than they select one unaided.
\item \textit{Robustness exposes instability.} Across systems, \passk{} spans
14--41\% agent-directed and 37--54\% scaffolded. Rankings can reverse: Gemini
beats GPT xhigh on regular \passone{} (51.5\% versus 44.5\%), yet trails on
\passk{} (13.5\% versus 20.0\%).
\item \textit{Diverse environments expose more than run-to-run variation.}
For scaffolded GPT xhigh, three regular runs have similar \passone{} rates
(74.5\%, 75.5\%, and 77.0\%), yet only 103 of 200 tasks pass
every repeat (\passk{} = 51.5\%): 86 tasks change outcome at least once.
Replacing two repeats with noise-heavy and speech-heavy conditions lowers the
all-three count to 77 (38.5\%).
Environment diversity therefore reveals instability beyond stochastic reruns,
and one-shot success overstates reliability.
\item \parbox[t]{\linewidth}{\textit{More reasoning does not consistently
improve robustness.} From minimal to xhigh, \passk{} rises from 15.0\% to 20.0\%
agent-directed but falls from 45.5\% to 38.5\% scaffolded.}
\end{list}

\subsection{Risk recognition and recovery}

\begin{figure}[t]
\centering
\includegraphics[width=\columnwidth]{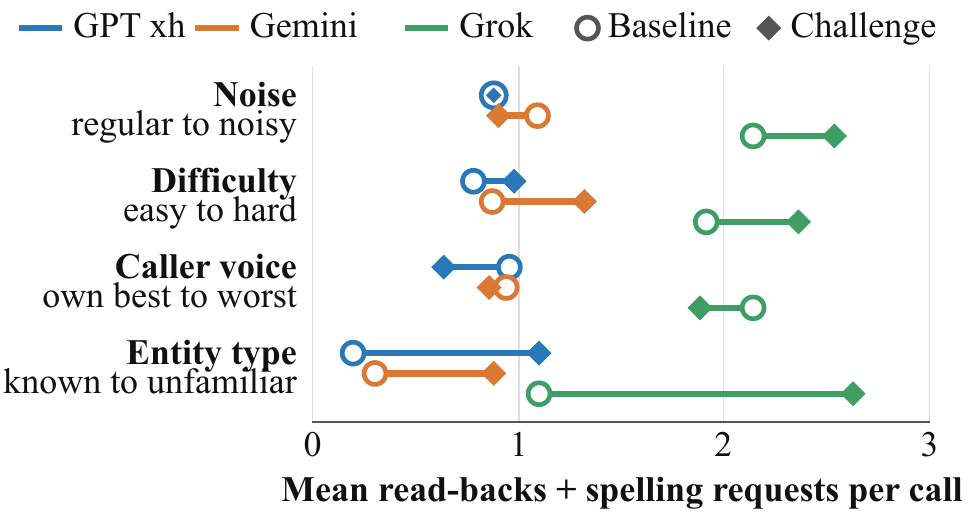}
\caption{Mean verification effort (read-backs plus spelling requests). All
systems increase effort for hard and unfamiliar entities, but not for their own
lowest-\passone{} caller voice; only Grok increases effort under noise.}
\label{fig:adaptivity-effort}
\end{figure}

Initial captures are wrong for 69\% of GPT xhigh fields, 65\% of Gemini fields,
and 42\% of Grok fields. GPT and Gemini verify 66\% and 64\% of wrong captures
versus 50\% of correct ones; Grok verifies nearly everything (94\%/96\%). The
systems likewise spend more effort on hard and unfamiliar entities, but not on
their weakest caller voice, and only Grok reacts to noise
(Figure~\ref{fig:adaptivity-effort}). Risk recognition is real but incomplete.

Only 27\%, 37\%, and 24\% of verified GPT xhigh, Gemini, and Grok errors end
correct. Among 347 initially wrong captures, recovery is 10\% without a repair
step, 26\% with read-back only, 24\% with spelling only, and 35\% with both.
Because agents choose when to verify, these are associations, but the pattern
locates the bottleneck after risk is noticed. Agents often double down on an
error or hallucinate a replacement rather than complete the repair.

Caller voice is another blind spot. In agent-directed regular calls, the gap
between each system's best- and worst-performing voices is 27 points for GPT
xhigh, 28 for Gemini, and 15 for Grok. The ordering is not stable: Mildred leads
GPT xhigh and Grok, whereas Wei leads Gemini. Only Gemini shows evidence that
success differs across all five voices after correction ($p=.040$); no
individual Gemini voice pair remains significant, so the result establishes
heterogeneity without isolating one contrast. Across systems, Mildred
significantly exceeds Priya, Mamadou, and Arjun, but not Wei. Yet no system
increases verification for its own lowest-performing voice, suggesting that
agents do not adapt their effort to voice-specific difficulty.

\subsection{Stress tests and diagnostics}

\begin{figure}[t]
\centering
\includegraphics[width=\columnwidth]{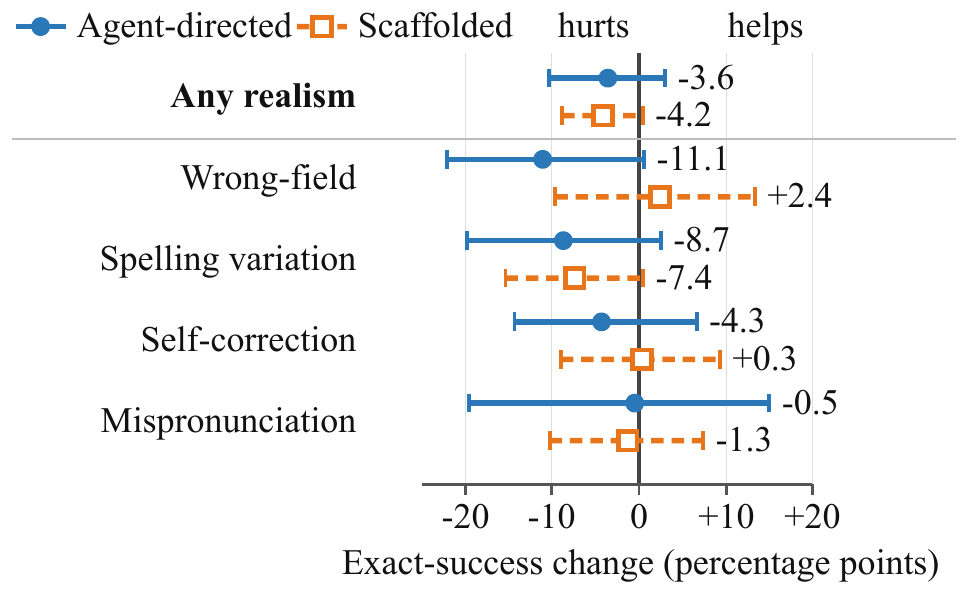}
\vspace{-22pt}
\caption{Caller-realism effects in agent-directed (solid circles) and
scaffolded (dashed squares) runs. Bars show task-clustered 95\% H\'ajek
intervals versus eligible clean assignments; none survives Holm correction.}
\label{fig:realism-effects}
\end{figure}

\noindent\textbf{Caller realisms mainly add repair cost.} No assigned realism
detectably changes exact success in either arm (Figure~\ref{fig:realism-effects}).
Overall H\'ajek estimates are $-3.6$ points (95\% CI $[-10.4,3.0]$)
agent-directed and $-4.2$ ($[-8.9,0.4]$) scaffolded. Assigned realisms may
remain latent. In agent-directed runs, mispronunciation changes success by only
$-0.5$ points but adds 24 points of spelling requests and 26 seconds. Realisms
therefore add repair cost without explaining the large baseline failure rate.

\begin{table}[t]
\caption{Value-bank construction and robust agent-directed results. Every bank
has 80 values and contributes 20 tasks (10 easy, 10 hard). \passk{} pools 60
system--task triples per bank.}
\label{tab:entity-results}
\centering
\small
\setlength{\tabcolsep}{2pt}
\renewcommand{\arraystretch}{.9}
\begin{tabular}{@{}>{\raggedright\arraybackslash}p{.25\columnwidth}
                    >{\raggedright\arraybackslash}p{.58\columnwidth}c@{}}
\toprule
Bank & Construction / grounding & \passk{} $\downarrow$ \\
\midrule
Dates        & Calendar and clock & .60 \\
Phones       & Regulator-reserved ranges & .47 \\
Times        & Calendar and clock & .43 \\
Person names & SSA + Census \cite{ssaNames,censusSurnames} & .30 \\
Codes        & ISO 3779 VINs; mod-97 IDs & .25 \\
Properties   & Registry-screened synthetic & .18 \\
Addresses    & Synthetic address grammar & .10 \\
Emails       & DNS-nonresolving domains & .08 \\
Coined       & Registry-screened synthetic & .03 \\
Medications  & RxNorm-checked \cite{nelson2011rxnorm} & .02 \\
\bottomrule
\end{tabular}
\end{table}

\smallskip
\noindent\textbf{Entity structure determines difficulty.} Across the regular
agent-directed cells, \passone{} is 64.0\% on easy entities and 44.3\% on hard
ones; scaffolding raises these rates to 87.3\% and 71.3\%. Pooled \passk{}
ranges from 1.7\% for medications and 3.3\% for coined names to 46.7\% for
phone numbers and 60.0\% for dates (Table~\ref{tab:entity-results}). The pattern
is consistent with format-constrained versus lexical capture. Dates and phone
numbers follow tightly scoped numeric schemas. Medications come from a closed
RxNorm vocabulary and coined names from a synthetic registry, but the agent is
shown neither list.

\begin{table}[t]
\caption{Effect of field validation and retry under the scaffolded prompt. Task
\passone{} requires every field to be correct; field \passone{} scores
individual values.}
\label{tab:protocol-ablation}
\centering
\small
\setlength{\tabcolsep}{4pt}
\renewcommand{\arraystretch}{.9}
\begin{tabular}{@{}>{\raggedright\arraybackslash}p{.51\columnwidth}cc@{}}
\toprule
Workflow & Task \passone{} & Field \passone{} \\
\midrule
Batch submit; no field validation & $.64\pm.06$ & $.78\pm.03$ \\
Validate each field; retry errors & $.82\pm.05$ & $.86\pm.03$ \\
\bottomrule
\end{tabular}
\end{table}

\begin{figure}[t]
\centering
\includegraphics[width=\columnwidth]{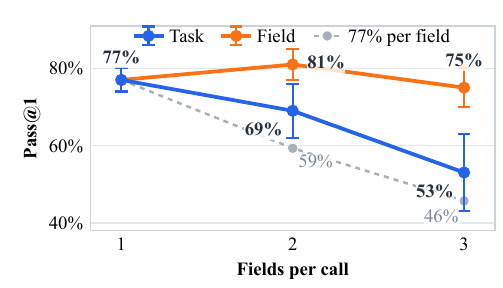}
\vspace{-16pt}
\caption{Task success declines as fields accumulate,
roughly following the light-gray reference expected if every field independently
succeeds 77\% of the time (77\%, 59\%, and 46\%). Error bars show 95\%
confidence intervals.}
\label{fig:composition-results}
\end{figure}

\smallskip
\noindent\textbf{Errors compound across fields.} At three fields, only 0.53 of
calls pass even though individual fields remain 0.75 accurate
(Figure~\ref{fig:composition-results}): one wrong field fails the entire call.
Intermediate validation and retry counter this accumulation, raising task
accuracy from 0.64 to 0.82 across the matched multi-field tasks
(Table~\ref{tab:protocol-ablation}). Because the intervention combines both
mechanisms, we do not attribute the gain to either one alone.

\begin{table}[!ht]
\caption{Share of scored agent utterances with a clear speech-fidelity defect
by system (lower is better).}
\label{tab:fidelity-provider}
\centering
\small
\setlength{\tabcolsep}{3pt}
\renewcommand{\arraystretch}{.9}
\begin{tabular}{@{}lcccc@{}}
\toprule
& GPT minimal & GPT xhigh & Gemini high & Grok \\
\midrule
Severity $\geq 2$ & 2.5\% & 2.1\% & 3.6\% & 4.1\% \\
\bottomrule
\end{tabular}
\end{table}

\smallskip
\noindent\textbf{Human review.} Across 90 failed calls, agreement or
adjudication assigns 81 failures to the agent and two to user-simulator logic;
no call is jointly attributed solely to infrastructure. Among the 81 agent
failures, agreed or adjudicated labels assign 42 as transcription, 16 as
logical, six as VAD, and two as hallucination; the remaining 15 retain subtype
disagreement. The speech-fidelity judge
compares agent audio with its reference transcript, flagging mispronunciations,
word additions, omissions, and clipping; severity $\geq 2$ marks a clear,
potentially confusing mismatch. The frozen set yields .80 precision, .75
recall, and .77 F1. A matched text run passes all 200
tasks, implicating spoken capture and repair rather than task understanding.

\section{Conclusion}

Voice agents still struggle to collect exact entities reliably. A strict
elicitation protocol helps substantially, and our caller is deliberately highly
cooperative: it spells accurately and corrects faulty read-backs. Even under
these favorable conditions, scaffolded systems succeed across all three
environments on only 37--54\% of tasks.

\vspace{3pt}
\noindent Failures occur at both stages of recovery: agents do not consistently
recognize risky captures, and when they do verify an error, only 24--37\% are
repaired.

\vspace{3pt}
\noindent\textbf{Limitations.} We study English, three providers, one simulator,
and one run per main environment. We intentionally omit caller-side ASR to
avoid confounding agent performance with a second recognizer. As a result,
agent speech synthesis is not evaluated end to end: the speech-fidelity judge
reports synthesis defects, but they do not yet affect task reward. The
multi-field study is formative; broader provider, human-caller, and
multilingual validation remain future work.

\vspace{\baselineskip}
\noindent\textbf{Acknowledgment.} We thank Ben Shi, Ola Zytek, Vijay Iyengar,
Ajeet Grewal, and Clay Bavor for feedback and support.

\clearpage

\section{Compliance with Ethical Standards}

The benchmark uses no real customer records or recordings. Values are
synthetic or public and non-sensitive; calls use simulated callers and provider
voices. Reviewers consented and were compensated. Before release, artifacts
will be screened for incidental personal data or secrets, and flagged items
will be excluded.

Exact-value elicitation is dual use: it can improve service calls but can also
facilitate collection of sensitive identifiers. Deployments should minimize
collection, explain its purpose, protect stored values, and support human
escalation. Our simulated English results are not evidence of safety, privacy,
deployment readiness, or differences among people or demographic groups.

This work was supported by Sierra and Mercor. The authors are employees of
Sierra and Mercor, respectively, and have no other relevant interests to
disclose.

\raggedcolsend
\bibliographystyle{IEEEbib}
\bibliography{references}

\end{document}